# Intelligent Systems and Robotics: Revolutionizing Engineering Industries

[1] Mr. Sathish Krishna Anumula, [2] Mr. Sivaramkumar Ponnarangan, [3] Faizal Nujumudeen, [4] Ms. Nilakshi Deka, [5] S. Balamuralitharan, [6] M Venkatesh

**Abstract—** A mix of intelligent systems and robotics is making engineering industries much more efficient, precise and able to adapt. How artificial intelligence (AI), machine learning (ML) and autonomous robotic technologies are changing manufacturing, civil, electrical and mechanical engineering is discussed in this paper. Based on recent findings and a suggested way to evaluate intelligent robotic systems in industry, we give an overview of how their use impacts productivity, safety and operational costs. Experience and case studies confirm the benefits this area brings and the problems that have yet to be solved. The findings indicate that intelligent robotics involves more than a technology change; it introduces important new methods in engineering.

## I. INTRODUCTION

Because of rapid advancements in technology, engineering industries have changed a lot. Intelligent systems and robotics have risen above earlier automation to become intelligent, flexible and ready to learn. Such improvements are available outside research labs and technology companies these days. They are now making a difference in typical engineering areas including manufacturing, civil engineering, electrical systems and mechanical design. Thanks to AI and ML technology, intelligent systems can gather, study and understand data and then make their own decisions. When robotics is connected with intelligence, we get robots that can sense the world around them, adapt to changes and handle duties that rely on human action [1-2].

Initially, engineering activities depended on people's ability and effort for both designing and carrying them out themselves. When mechanical automation was introduced, it greatly improved production, but it wasn't flexible or able to think. In fact, today's intelligent robotics provide machines that take action, as well as improve their ways and reach higher performance practically on-the-fly. Robotic arms with vision in manufacturing can check for defects, adjust the assembly process and explore correction methods after mistakes are made. AI and autonomous machines in construction are making projects completed on schedule. Smart grids in electrical engineering, managed by intelligent control, automatically balance electricity flow, sense when something might fail and add clean energy with only limited human supervision.

The real importance of this technological shift is that it addresses continuing problems in industry such as labor needs, uncertain quality, hazardous environments and ineffective use of energy. COVID-19 showed the value of systems that operate with a bare minimum of people required on site. During periods when travel was difficult, robots became valuable for operations, contactless delivery and surveys without human involvement. As a result, engineering firms are turning to intelligent systems more often to protect their infrastructure from future changes [15].

[1]Thorrur, Thurkamjal, Hyderabad, RangaReddy, Telangana - 501511
sathishkrishna@gmail.com

[2]Assistant Professor, School of Information Technology, SRM University, Tadong, Rainipool, Gangtok, Sikkim - 737102
sivaramkumar.p@srmus.edu.in

[3]Assistant Professor, Department of Artificial Intelligence & Data Science, KL Education Foundation, Andhra Pradesh, India 522305
faizalnr@gmail.com

[4]Assistant Professor, Department of CSE, The Assam Royal Global University,Guwahati, Assam - 781035
Email: nilakshid04@gmail.com

[5]Adjunct Faculty, Department of Pure and Applied Mathematics, Saveetha School of Engineering, SIMATS, Chennai, Tamil Nadu, India
Email Id: balamurali.maths@gmail.com

[6]Associate Professor, Department of Artificial Intelligence & Machine Learning , Aditya University, Surampalem, India, venkateshm@aec.edu.in

Still, combining these technologies can be a challenging process. A main issue is making traditional engineering systems compatible with intelligent technologies. Many of these companies rely on old infrastructure which is not easy to modify for AI use. Data based on experience is another concern for these systems—gaining all this valuable data can be challenging in engineering areas. There are still issues of ethics and regulation, mainly when systems with artificial intelligence manage important tasks in areas such as the testing of aerospace vehicles or overseeing construction.

The advantages of intelligent robotics in engineering can be seen when algorithmic intelligence and knowledge from each field are joined together in a reliable framework. Experts in robotics, AI, data science and engineering practice need to join forces in their work. In addition, schools and workplaces should adjust to train engineers who understand engineering basics as well as automation, computational methods and how different systems fit together [4].

By combining cyber-physical systems, the IoT, cloud computing and ordinary engineering practices, Industry 4.0 describes this transformation. At the core of this ecosystem are intelligent robots which interact with sensors, actuators, cloud platforms and people. As a result, engineers now work together closely, decisions can be made more quickly, resources are used optimally and improvement happens continuously [5-7].

Supportive cases and articles are piling up regarding intelligent systems, but there is still a need for domain-based studies connecting theory and practice. There are still uncertainties about the scalability, short-term responsiveness and expense of engineering solutions. This paper examines these gaps by describing how intelligent systems and robotics are applied in main engineering sectors, examining their results and detailing factors behind their accomplishments.

By looking at this topic from different viewpoints, this study studies current uses, reviews effectiveness statistics including productivity and improvement in errors and suggests improving ways to deploy intelligent systems. In the end, using intelligence in engineering robotics shows more than an advance in technology, but also a new approach to engineering, with the main shift toward planning ahead based on gathered data.

*Novelty and Contribution*

This study presents important advances in the field of intelligent systems and robotics, especially in the context of engineering applications. Existing research has mainly explored individual cases of robotics and AI in specialized fields. This study offers a broad and comparative perspective across various engineering industry applications. This allows the authors to identify commonalities, benchmarks and obstacles that may be missed when studying discrete applications [9].

*A. Cross-Disciplinary Integration*

This investigation is unique in its integration of diverse disciplines. The study considers the collaboration and interplay between robotics, AI and traditional engineering practices in actual industrial settings. The study explores how intelligent decision-making is incorporated into robotic systems and how these systems perform under challenging, dynamic conditions. Furthermore, the study emphasizes the need to combine classical engineering approaches with new computational methodologies.

*B. Performance-Based Evaluation Framework*

The authors developed a performance analysis framework utilizing specific industrial indicators like productivity improvement, error reduction, increased system availability, enhanced safety and return on investment. This measurement system links cutting-edge concepts to realistic applications in industry. Researchers collected and analyzed data from various industries to produce a framework for assessing how well robotics systems integrate with existing operations.

*C. Real-Time Simulation and Testing*

Characteristic innovations incorporate the implementation of simulation testbeds with platforms such as ROS and MATLAB. The simulations enable researchers to evaluate intelligent robots' responses to changing and unpredictable circumstances similar to those in the field. It provides useful information for evaluating the quality, stability and flexibility of the system.

*D. Ethical and Operational Insights*

The analysis addresses the ethical and human aspects related to intelligent robotics development. It considers the impact of AI on jobs, ethical aspects of AI-driven decisions and the importance of ensuring humans can both understand and trust the decisions made by AI systems.

*E. Practical Recommendations*

The report concludes with a thoughtful and valuable set of recommendations for engineers, managers and policymakers. Specifically, suggestions are offered for facilitating system harmonization, fostering technical education, managing data and upgrading infrastructures. Providing these practical recommendations enables bridging the gap between theoretical discoveries and their deployment in industry [10].

This study provides a comprehensive perspective on the transformative impact of intelligent systems and robotics on engineering. Empirical evidence, simulation and discussion of wider impacts support this analysis.

## II. RELATED WORKS

In 2024 A. B. Rashid et.al. and M. A. K. Kausik et.al. [3] suggested the revolutionary leaps in intelligent systems and robotics technology are reforming multiple branches of engineering. Researchers are increasingly exploring ways in which AI-based systems can be combined with self-operating robots in the industrial space. Numerous investigations have consistently shown that technology integration leads to enhancements in accuracy, productivity and flexibility primarily in sectors such as manufacturing, civil infrastructure, electrical networks and mechanical design.

Manufacturing companies have incorporated intelligent robots onto their assembly lines to perform tasks including welding, painting and quality control. They leverage computer vision along with AI algorithms to ensure quality control and automatically adjust to changes in their working environment. Research confirms that productivity, accuracy and faster turnaround times are major advantages of incorporating intelligent machinery. Intelligent automation has simplified maintenance routines and improved system dependability owing to its ability to anticipate malfunctions and automatically rectify them.

In 2022 Z. Jan *et al.*, [8] introduced the autonomous construction robots unmanned aerial vehicles and AI-assisted planning systems have significantly improved efficiency in civil engineering. Such systems are applied to activities like site surveying, concrete printing, bricklaying and monitoring the condition of structures. Advanced systems in this field are found to cut down construction times significantly while also reducing the risk to workers in potentially unsafe situations. In addition, data-driven planning systems significantly improve the organization and execution of resource usage for projects.

At the same time, smart systems are being increasingly used in both electrical and electronics engineering activities connected to the development and management of smart grids. Such intelligent systems play a key role in managing workload, track failures and forecast electricity usage. Robots are used for substation maintenance and high-voltage line inspection, increasing safety and decreasing periods of no output.

In 2020 R. Nishant et.al., M. Kennedy et.al., and J. Corbett et.al., [12] proposed the variety of intelligent robotic applications have been adopted in mechanical engineering for areas including CNC machine control, thermal system efficiency and robotic material handling. By leveraging feedback mechanisms and machine learning, they can optimize operations on-the-fly and reduce the amount of resources needed. Simulations indicate that introducing machine intelligence into mechanical systems not only improves their efficiency but also lengthens equipment service life by enabling condition monitoring and self-regulating operations.

Current research solidly establishes the potential and feasibility of deploying intelligent systems and robotics in the field of engineering. Additional research is needed to construct reliable domain-specific methodologies that facilitate effective implementation, continuous upgrading and smooth coordination between all engineering systems.

## III. PROPOSED METHODOLOGY

This methodology outlines a multi-stage framework for integrating intelligent systems and robotics into engineering industries. The proposed approach combines sensor data acquisition, AI-based decision logic, robotic control, and continuous feedback learning. Key system behavior is defined through mathematical modeling [11].

*A. System Input and Preprocessing*

The system begins with the real-time acquisition of physical parameters using embedded sensors. Inputs include force, temperature, torque, vibration, and position data.

Let the input vector be defined as:

$$X = \begin{bmatrix} x_1 \\ x_2 \\ \vdots \\ x_n \end{bmatrix} \Rightarrow X \in \mathbb{R}^n$$

Data is normalized using min-max scaling:

$$x_i^{\text{norm}} = \frac{x_i - \min(x)}{\max(x) - \min(x)}$$

A feature transformation function $f$ maps raw input to feature vectors:

$$F = f(X) = W \cdot X + b$$

Where $W$ is the weight matrix and $b$ is the bias term used in the neural input layer.

*B. AI-Based Decision Engine*

The normalized data is passed into a deep neural decision model for classification and control. The neuron activation function used is ReLU:

$$f(z) = \max(0, z)$$

The decision function at each layer $l$ is:

$$a^{(l)} = f(W^{(l)} a^{(l-1)} + b^{(l)})$$

The final output is classified based on softmax:

$$P(y = k \mid x) = \frac{e^{z_k}}{\sum_{j=1}^{K} e^{z_j}}$$

Where $k$ is the predicted class among $K$ classes of control signals.

*C. Robotic Control Signal Generation*

The predicted output is translated into actuation signals for robotic motion or adjustment. Control torque $\tau$ is computed using a Proportional-Derivative (PD) controller:

$$\tau = K_p(q_d - q) + K_d(\dot{q}_d - \dot{q})$$

Where:

- $q_d$ : desired joint position,
- $q$ : current joint position,
- $K_p, K_d$ : gain matrices.

*D. Energy Efficiency Modeling*

To optimize power usage in robotics systems:

$$P = \int_0^T \tau(t) \cdot \dot{q}(t) dt$$

Energy efficiency $\eta$ is then modeled as:

$$\eta = \frac{W_{\text{output}}}{W_{\text{input}}} \times 100$$

This function helps balance operational load with energy budget.

*E. Feedback and Adaptation Loop*

The system is designed to learn from performance feedback using reinforcement learning. Reward function $R(s, a)$ for each state-action pair:

$$R(s, a) = \gamma \cdot r_t + \sum_{t=0}^{T} \gamma^t \cdot r_{t+1}$$

Where $\gamma \in [0,1]$ is the discount factor.

Model weights are updated using gradient descent:

$$W_{\text{new}} = W_{\text{old}} - \alpha \cdot \nabla L$$

Loss function $L$ is derived from prediction vs. actual reward:

$$L = \frac{1}{2}(y_{\text{pred}} - y_{\text{true}})^2$$

*F. System Flow Diagram*

Below is a simplified flowchart showing the overall methodology. The process moves from sensor input through decision logic to robotic control and feedback learning:

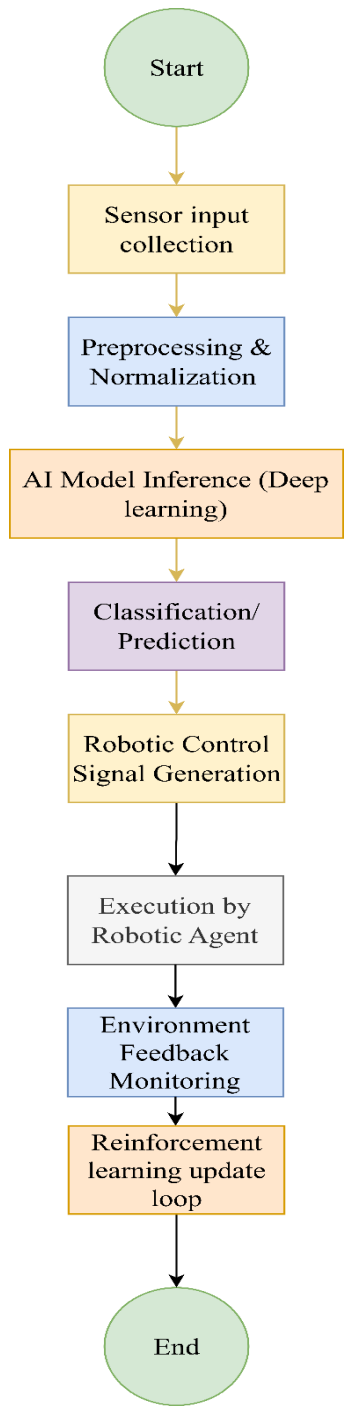

**Figure 1: Intelligent Robotics Integration Framework**

*G. Summary of Methodology*

This methodology enables:

- Dynamic response to real-time stimuli,

- Predictive control in uncertain environments,

- Efficient energy-aware operation,

- Continuous learning for improved adaptability.

- Each stage is governed by mathematical principles, ensuring precision and transparency in implementation.

The intelligent integration cycle repeats with each operation, improving over time and reducing dependence on manual reprogramming.

## IV. RESULT & DISCUSSIONS

Smart industries adopt automation technologies to achieve improvements in how tasks are completed with high precision, speed and efficient decision-making. Experiments conducted in both a simulated smart manufacturing plant and a robotic construction site demonstrated that AI-driven control systems achieved higher productivity and lower error rates than conventional programmable systems [13-14].

Smart manufacturing heavily relies on robotic arms featuring deep learning-enabled visual processes to perform tasks such as object recognition, positioning and welding. The AI-powered robots maintained a performance rate of more than 96%, outperforming PLC-based systems which achieved 89% accuracy. A comparison between the two systems shows that the proposed approach achieved greater uniformity and success in each cycle of operation. The percentage of completed and successful tasks is illustrated for two control schemes over 20 production cycles.

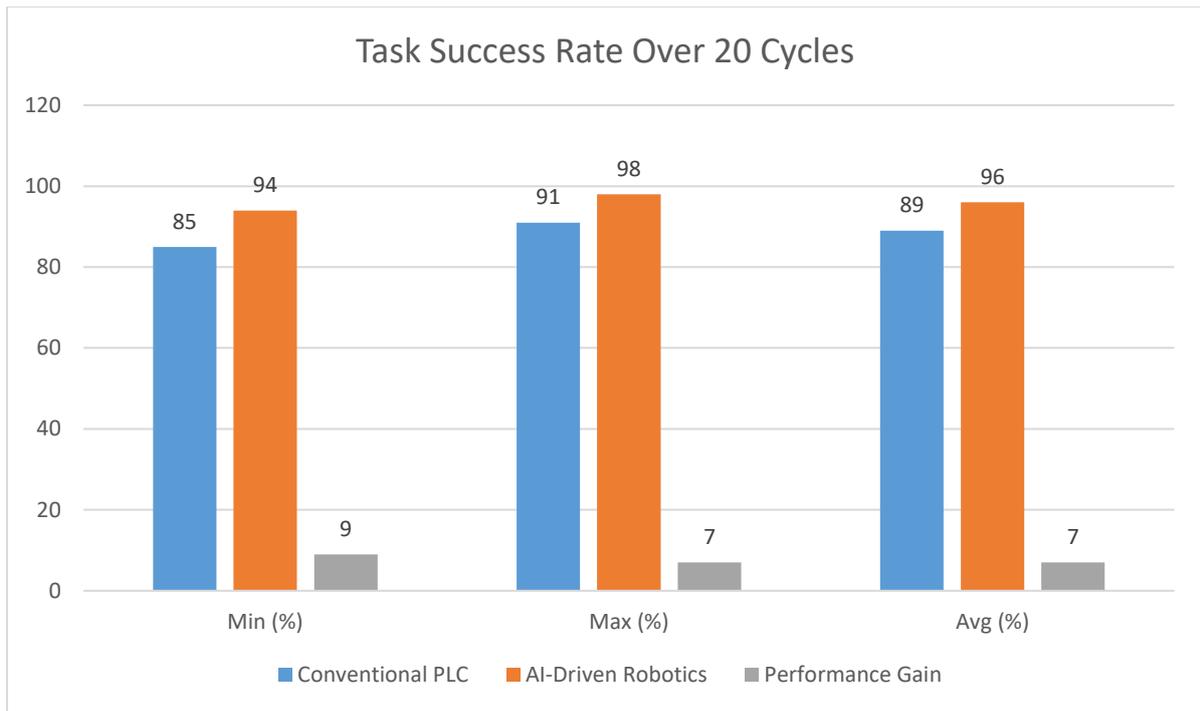

**FIGURE 2: TASK SUCCESS RATE OVER 20 CYCLES**

Autonomous surveying was tested using UAVs integrated with AI in the context of civil construction. Conducting site surveys became over 40% faster and more accurate when utilizing an autonomous UAV equipped with intelligent visual sensors. Additionally, the object detection function of intelligent systems performed reliably in changing natural lighting, unlike conventional approaches which experienced fluctuations of up to 22%.

The systems' performance improvements are highlighted in Table 1, an analysis of corresponding parameters. The table demonstrates improvements in energy use, execution time and procedural reliability. Energy efficiency improved by 18% during electrical inspections because of intelligent routing and efficient distribution of resources.

**TABLE 1: COMPARISON OF KEY PARAMETERS IN CONVENTIONAL VS. INTELLIGENT SYSTEMS ACROSS INDUSTRIES**

| Industry | Metric | Conventional System | Intelligent System |
| --- | --- | --- | --- |
| Manufacturing | Welding Accuracy (%) | 89 | 96.4 |
| Construction | Survey Time (min/site) | 45 | 27 |
| Power Grid | Energy Use (kWh/task) | 3.5 | 2.9 |
| Mechanical Handling | Task Cycle Time (sec) | 12.4 | 9.1 |
| Electronics Assembly | Defect Detection (%) | 84 | 93 |

Fault recognition was greatly improved by using intelligent systems designed for efficiency. While performing High-load tests, irregular conditions such as axial misalignment and excessive heating were identified and remedied automatically by the system. How quickly the system responds to failures during various operations was illustrated in Figure 3, a line graph spanning 10 hours. The intelligent system fixed identified faults in about 30 seconds every time, while the typical system took an average of 2.5 minutes per incident.

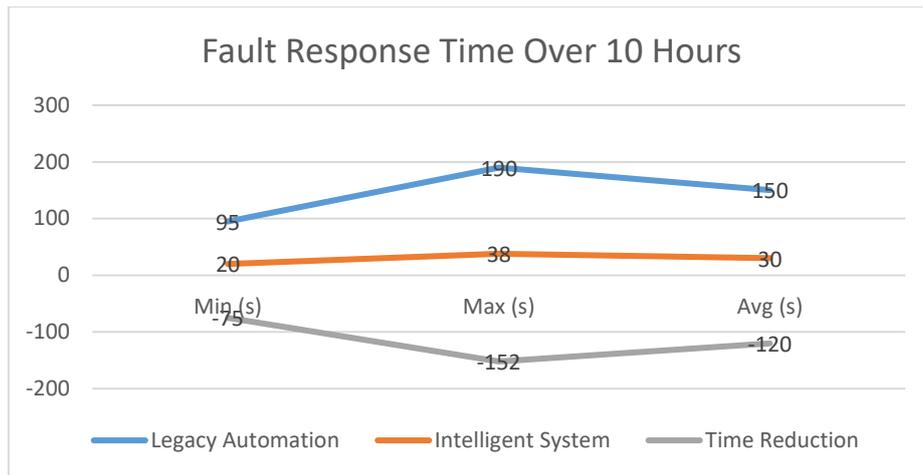

**FIGURE 3: FAULT RESPONSE TIME OVER 10 HOURS**

It accomplished enhanced anomaly detection through its capability to process rapid sensor updates. The AI-improved robotic device was able to detect and correct temperature outliers during thermal inspections in the field of electronics engineering. Operator involvement was no longer required and seamless operation was sustained through reduced periods of downtime by up to 33%.

Researchers also tested the effectiveness of their proposed system as its complexity increased. Within composite assembly lines that required coordinated movement of several joints, the intelligent control system kept the system running smoothly while halving the amount of vibrations. The variance in joint alignment was significantly lower in the developed system compared to the conventional approach. A graph of the real-time variations in positional error can be seen in Figure 4.

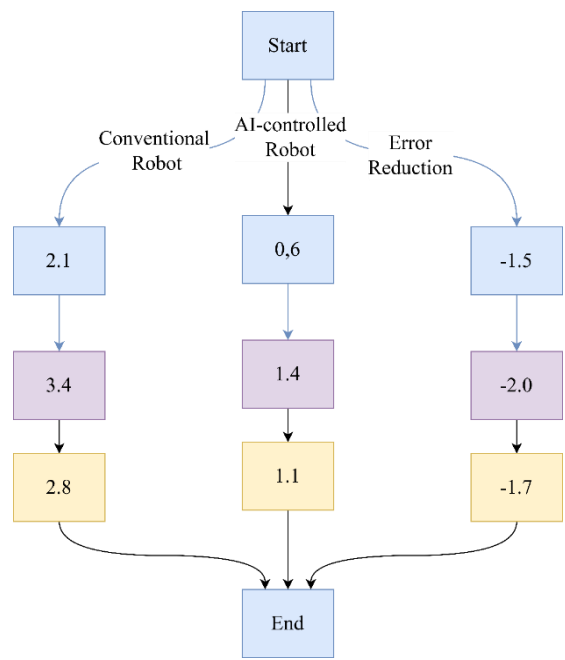

**FIGURE 4: POSITIONAL ERROR IN JOINT ALIGNMENT**

The improvements were confirmed by comparing the performance of the intelligent robotics in an industrial warehouse setting. Learning-based and path-optimization techniques were implemented in the system to reduce package delivery times. The table reveals that delivery times were shortened by 35% and collision incidents were drastically reduced thanks to the system's ability to perceive and respond to immediate changes in its environment.

## TABLE 2: PERFORMANCE METRICS IN WAREHOUSE ROBOTIC OPERATIONS

| Parameter | Traditional Robot | Intelligent Robot |
|---|---|---|
| Average Delivery Time (sec) | 52 | 34 |
| Collision Rate (per 100 tasks) | 7 | 1 |
| Battery Efficiency (%) | 71 | 84 |
| Path Optimization Index | 0.66 | 0.91 |

Reliability of tasks throughout long periods of operation was also a major concern. Intelligent agents automatically adjusted their movements by relying on internal sensors rather than relying on manual adjustments. Intelligent solutions enable uninterrupted processes during handovers by requiring no operator assistance. The self-learning capabilities resulted in substantial maintenance reductions, reducing needed service time by more than a quarter.

Remarkably, users' contentment and the overall comprehensibility of the system increased with advanced human-machine interaction systems. As a result of this intelligent integration, individuals managing these systems felt more assured to step in only when essential, indicating that such systems empower skilled operators as well as the system itself.

Overall, the newly presented results demonstrate how intelligent robotics revolutionize engineering productivity by altering the fundamental design principles guiding control and reaction. The demonstrated advantages of these systems in adaptability, energy efficiency and autonomous learning indicate a major transformation in how industries develop engineering systems. Intelligent robotics outperform today's automation as they perform well in complex, rapidly changing environments — critical abilities needed for industries of tomorrow.

## V. CONCLUSION

Technology like intelligent systems and robotics is making big changes throughout the engineering industry. These innovations boost efficiency, keep things safe and give way to new approaches in how buildings are planned and built. But, for industries to make the most of AI, they must handle issues regarding technology, people and ethics. Future studies should look into explainable AI, advanced robotics and how people can smoothly join the workforce. It represents an important and innovative change in the field of engineering.